\documentclass{article}
\usepackage{ijcai25}

\usepackage{times}
\usepackage{soul}
\usepackage{url}
\usepackage[hidelinks]{hyperref}
\usepackage[utf8]{inputenc}
\usepackage[small]{caption}
\usepackage{graphicx}
\usepackage{amsmath}
\usepackage{amsthm}
\usepackage{booktabs}
\usepackage{algorithm}
\usepackage{algorithmic}
\usepackage[switch]{lineno}
\usepackage{subfig}

\title{Rethinking Contrastive Learning in Graph Anomaly Detection: A Clean-View Perspective}

\author{
    Author Name
    \affiliations
    Affiliation
    \emails
    email@example.com
}

\author{
Di Jin$^1$
\and
Jingyi Cao$^1$\and
Xiaobao Wang$^{1,2,}$\thanks{Corresponding author}\and
Bingdao Feng$^1$ \and
Dongxiao He$^1$  \and
Longbiao Wang$^1$\And 
Jianwu Dang$^3$\\
\affiliations
$^1$College of Intelligence and Computing, Tianjin University, Tianjin, China\\
$^2$Guangdong Laboratory of Artificial Intelligence and Digital Economy (SZ), Shenzhen, China\\
$^3$Shenzhen Institute of Advanced Technology, Chinese Academy of Sciences, Shenzhen, China\\
\emails
\{jindi, caojingyi, wangxiaobao, fengbingdao, hedongxiao, longbiao$\_$wang\}@tju.edu.cn,
jdang@jaist.ac.jp
}

\begin{document}

\maketitle

\begin{abstract}
    Graph anomaly detection aims to identify unusual patterns in graph-based data, with wide applications in fields such as web security and financial fraud detection. Existing methods typically rely on contrastive learning, assuming that a lower similarity between a node and its local subgraph indicates abnormality. However, these approaches overlook a crucial limitation: the presence of interfering edges invalidates this assumption, since it introduces disruptive noise that compromises the contrastive learning process. Consequently, this limitation impairs the ability to effectively learn meaningful representations of normal patterns, leading to suboptimal detection performance. To address this issue, we propose a Clean-View Enhanced Graph Anomaly Detection framework (CVGAD), which includes a multi-scale anomaly awareness module to identify key sources of interference in the contrastive learning process. Moreover, to mitigate bias from the one-step edge removal process, we introduce a novel progressive purification module. This module incrementally refines the graph by iteratively identifying and removing interfering edges, thereby enhancing model performance. Extensive experiments on five benchmark datasets validate the effectiveness of our approach.
\end{abstract}

\section{Introduction}

Graph, a fundamental data structure comprising nodes and edges, plays a crucial role in representing relationships across diverse disciplines, such as recommendation systems~\cite{wu2019session,he2024tut4crs}, backdoor attack~\cite{feng2024backdoor,jin2025backdoor}, and image recognition~\cite{liu2024hierarchical,du2023superdisco}. Within graph analytics, Graph Anomaly Detection (GAD) has emerged as a critical area of research~\cite{XiangZHQDDB023,xiang2024federated}. It aims to identify patterns that significantly deviate from the majority of instances~\cite{noble2003graph,kim2022graph}. The detection of such anomalies reveals latent irregularities in the data, allowing proactive interventions to safeguard data integrity. This capability has far-reaching applications, particularly in fields such as sarcasm detection~\cite{wang2023augmenting,wang2025elevating}, defense against attack~\cite{jin2023local}, and the discovery of pathological mechanisms in the brain~\cite{Lanciano2020ExplainableCO}.

Early methods employ shallow mechanisms such as ego-network analysis~\cite{perozzi2016scalable}, residual analysis~\cite{li2017radar}, and CUR decomposition~\cite{peng2018anomalous} to detect anomalies. However, these approaches fail to capture the complex relationships inherent in graph data, which limits their ability to detect sophisticated anomalies. With the rise of deep learning, some studies~\cite{ding2019deep,fan2020anomalydae,luo2022comga} utilize Graph Neural Networks (GNNs) to reconstruct structures and node features. They use reconstruction errors as a basis for anomaly identification. Despite their advances, these methods demand extensive memory resources. In addition, the convolutional operations in GNNs can smooth out anomalous signals, thereby diminishing the distinctiveness of anomalous nodes and affecting detection accuracy. More recently, researchers have employed contrastive learning to tackle the challenges of GAD~\cite{liu2021anomaly,jin2021anemone,zhang2022reconstruction}. These methods utilize Random Walk With Restart (RWR)~\cite{tong2006fast} to sample subgraphs, generating positive instance pairs between nodes and their local subgraphs, as well as negative instance pairs from different subgraphs. Anomalies are detected by comparing the similarities between these instance pairs, which prevents the smoothing of anomalous signals.

\begin{figure*}
    \centering
    \includegraphics[width=1.0\linewidth]{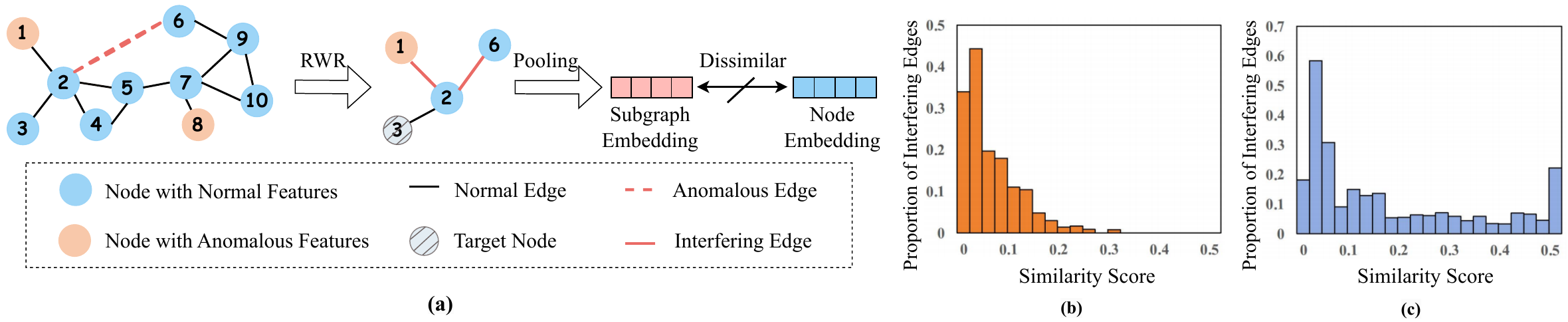}
    \caption{(a) A toy example illustrating GAD problems based on contrastive learning. In the whole graph, the red dashed line represents an anomalous edge—a connection that did not originally exist between nodes. The red node signifies an anomalous node, whose features are altered. Both of these anomalies introduce interfering edges (shown as red solid lines), which disrupt the process of generating positive instance pairs. The middle subgraph is sampled through RWR starting from node 3, after which the subgraph embedding for node 3 is obtained through further processing; (b) The proportion of interfering edges under different similarity scores based on raw features on the Cora dataset; (c) The proportion of interfering edges under different similarity scores based on GCN-aggregated features on the Cora dataset.}

    \label{fig:motivation}
\end{figure*}

However, existing contrastive learning-based methods have a significant flaw: \textbf{interfering edges} undermine the fundamental premise of contrastive learning, which relies on the similarity between instance pairs for effective model training. This issue stems from the presence of anomalies in graphs—contextual and structural anomalies~\cite{liu2021anomaly,duan2023normality}. Contextual anomalies manifest through alterations in node features. These distortions are propagated through edges connected to nodes with anomalous features, which disrupt the semantic consistency of the subgraph. Structural anomalies, on the other hand, occur when anomalous edges link previously unconnected nodes, altering the subgraph’s inherent topology. Since both types of edges inevitably disrupt the process of generating positive instance pairs, we define them as ‘\textbf{interfering edges}.' As depicted in Figure \ref{fig:motivation}(a), if node 3 samples nodes 2, 1, and 6 to form a subgraph, the model may mistakenly consider the instance pair formed by node 3 and the subgraph comprising nodes 1, 2, and 6 as a positive instance pair, incorrectly perceiving them as similar. This distortion in positive instance similarity impairs the model’s ability to effectively learn normal patterns, ultimately reducing its capacity to distinguish between normal and anomalous data. 

To mitigate the negative impact of interfering edges on contrastive learning, it is essential to accurately identify and remove these edges during the training process.  A straightforward approach is to remove edges with low raw feature similarity~\cite{Aghabozorgi2018ANS,KUMAR2020124289}. However, as depicted in Figure \ref{fig:motivation}(b), relying solely on raw feature similarities is insufficient. Although such a method seems intuitive, it is inherently myopic—focusing solely on feature-level discrepancies while neglecting the multi-dimensional structure of the graph. To address this limitation, we also explore an edge removal method based on the similarity of GCN-aggregated embeddings, which is depicted in Figure \ref{fig:motivation}(c). Unfortunately, the results remain unsatisfactory due to the inherent tendency of GCNs to smooth node representations, which artificially increases the feature similarity between connected nodes, including those that are anomalous. The core challenge, therefore, lies in effectively integrating both feature and structural information to precisely identify interfering edges. Furthermore, a one-step edge removal strategy often leads to incomplete or biased elimination of interfering edges. During the initial phase of training, CVGAD is still susceptible to these edges, which hinders its ability to accurately assess the degree of interference. Thus, another critical challenge is to minimize the impact of interfering edges during the training phase to enhance overall performance.

In this paper, we propose a \(\mathbf{C}\)lean-\(\mathbf{V}\)iew enhanced \(\mathbf{G}\)raph \(\mathbf{A}\)nomaly \(\mathbf{D}\)etection model (\(\mathbf{CVGAD}\)). Our approach incorporates a multi-scale anomaly awareness module that employs a dual-scale contrastive learning framework, operating simultaneously on both anomalous and clean graphs. By leveraging node-subgraph (NS) contrast and node-node (NN) contrast, the module effectively integrates feature and structural information to accurately evaluate the degree of anomalies. These scores underpin the construction of an interference-sensitive edge detection matrix, which identifies edges that introduce noise and disrupt the training process. Unlike the one-step edge removal method, CVGAD employs a novel progressive purification module that incrementally refines the graph by iteratively removing edges with high interference scores. These scores are dynamically recalculated at each step using continuously updated node scores, gradually minimizing the effect of interfering edges during the training phase. The synergy between the multi-scale anomaly awareness module and the progressive purification module significantly enhances the model’s overall performance.

In summary, our contributions are as follows:
\begin{itemize}
\item We discover a fundamental flaw in contrastive learning-based GAD methods: interfering edges undermine the core principles of contrastive learning.
\item We propose a novel method that iteratively identifies and removes interfering edges to obtain a clean graph, mitigating their adverse impact on contrastive learning.
\item Through extensive experiments on five benchmark datasets, we validate the effectiveness of our edge removal strategy and highlight CVGAD's superior performance compared to baselines.
\end{itemize}

\begin{figure*}
    \centering
    \includegraphics[width=0.85\linewidth]{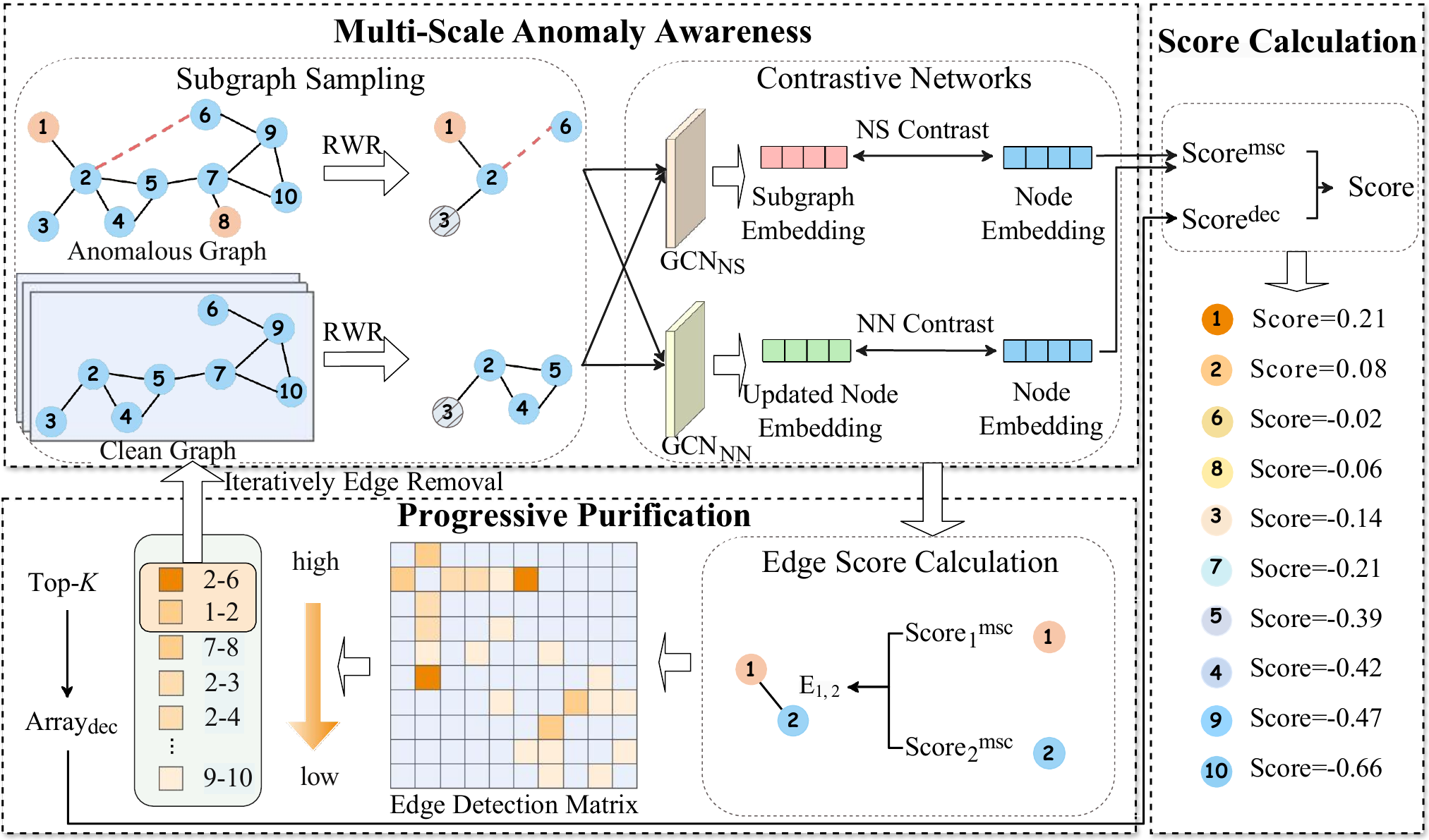}
    \caption{Overview of CVGAD. It consists of three primary components: (1) Multi-scale anomaly awareness: NS contrast and NN contrast are conducted on both the anomalous graph and clean graph, jointly training the model; (2) Progressive purification: Node contrast scores are calculated on the anomalous graph. Based on these scores, an interference-sensitive edge detection matrix is generated. The edges with the top-\textit{K} highest interference scores are removed to create a cleaner graph. This module, together with the multi-scale anomaly awareness component, is trained iteratively to incrementally improve model performance; (3) Score calculation: The contrast score and the detection score are combined to generate the final node anomaly value.}
    \label{fig:framework}
\end{figure*}

\section{Related Work}

Graph anomaly detection aims to identify abnormal patterns that significantly deviate from typical patterns~\cite{noble2003graph,kim2022graph}. Early methods mainly focus on shallow techniques. AMEN~\cite{perozzi2016scalable} leverages the ego-network information of each node to identify anomalies. Radar~\cite{li2017radar} identifies anomalies through the residual of features and network information. ANOMALOUS~\cite{peng2018anomalous} utilizes CUR decomposition and residual analysis within attributed networks. However, shallow methods have limitations in effectively handling complex patterns and learning intricate relationships. Conversely, deep learning exhibits a superior capacity for non-linear modeling and showcases its proficiency in addressing these complexities~\cite{hamilton2017inductive}. Dominant~\cite{ding2019deep} and ADA-GAD~\cite{he2024ada} employ GNNs to reconstruct the graph topology and node features, identifying nodes with large reconstruction errors as anomalies. HCM~\cite{huang2022hop} predicts the path lengths between nodes and regards hop counts as anomaly indicators. SI-HGAD~\cite{zou2024structural} combines hierarchical information and conducts hierarchical substructural modeling. CoLA~\cite{liu2021anomaly} implements contrastive learning in graph anomaly detection for the first time. It compares the similarities between nodes and their local subgraphs to identify anomalies. ANEMONE~\cite{jin2021anemone} further incorporates node-node comparisons and detects anomalies through multi-scale contrast. SL-GAD~\cite{zheng2021generative} leverages a generative attribute regression module combined with a contrastive learning framework for anomaly detection. Sub-CR~\cite{zhang2022reconstruction} generates a new view using graph diffusion algorithms, conducting contrastive learning and attribute reconstruction on both views to obtain anomaly scores. NLGAD~\cite{duan2023normality} improves the detection performance by means of the refined training on normal nodes. GRADATE~\cite{duan2023graph} compares the original view with the augmented view, enhancing robustness through multi-view contrast. DiffGAD~\cite{li2024diffgad} further employs the diffusion model in graph anomaly detection and proposes a DM-based detector. However, these methods overlook the impact of interfering edges on contrastive learning, which disrupt the training process and degrade model performance.

\section{Method}
\subsection{Notations}
An attributed graph is represented as \(\mathcal{G}=(\mathcal{V}, \mathcal{E})\), where \(\mathcal{V}=\left\{v_{1}, v_{2}, \ldots, v_{n}\right\}\) denotes the set of nodes and \(\mathcal{E}\) denotes the set of edges. The adjacency matrix \(\mathbf{A}\in\mathbf{R}^{n \times n}\) provides the structure information within the graph. Specifically, \(\mathbf{A}_{i, j}=1\) indicates that there exists an edge between node \(v_{i}\) and \(v_{j}\); otherwise, \(\mathbf{A}_{i, j}=0\). The matrix \(\mathbf{X}\in\mathbf{R}^{n \times o}\) contains information about node features. In anomaly detection, the objective is to learn a function 
\(f{(\cdot)}\) to evaluate the anomaly score for each node in the graph. The higher the score, the more likely the node is to be considered an anomaly.

\subsection{Framework Overview}
The overall pipeline of CVGAD is illustrated in Figure \ref{fig:framework}. It consists of three interconnected modules. First, the multi-scale anomaly awareness module utilizes RWR~\cite{tong2006fast} to extract subgraphs from dual views. Discriminators are subsequently employed for node-subgraph contrast and node-node contrast. This module evaluates the complex relationships between nodes and their local subgraphs. Next, the progressive purification module applies specific rules to calculate the contrast scores for nodes and the interference scores for edges. By analyzing and filtering these interference scores, edges with higher scores are selected and removed to generate a clean graph. This process is iterated to achieve optimal results. Finally, the score calculation module integrates the contrast scores with the detection scores to derive the final anomaly values. In the following sections, we will provide detailed descriptions of these modules.

\subsection{Multi-Scale Anomaly Awareness}
\textbf{Subgraph Sampling.} The anomaly level of a node correlates intricately with the structure and features of its local subgraph. Normal nodes typically share high similarities with neighboring nodes, while anomalies exhibit the opposite behavior. Therefore, we employ RWR~\cite{tong2006fast} to sample subgraphs \(\mathcal{G}^{a}\) and \(\mathcal{G}^{c}\) from the anomalous and clean graphs, respectively. Each subgraph includes the target node and neighboring nodes, with a fixed size of \(N\). It is worth noting that the clean graph remains consistent with the anomalous graph in the initial phase. On the one hand, RWR on the clean graph minimizes the impact of interfering edges on contrastive learning. On the other hand, RWR on the anomalous graph encourages the model to learn rich and diverse features. Additionally, to avoid the model's easy recognition of the target node in the subgraph, we mask the information of the target node. Nodes and masked subgraphs serve as the inputs for node-subgraph contrast and node-node contrast.

\textbf{Node-Subgraph Contrast.} The objective of node-subgraph contrast is to learn the coherence between a node and its entire local subgraph~\cite{liu2021anomaly}. For this purpose, we construct instance pairs by combining node embeddings with masked subgraph embeddings. Since raw features typically only represent the characteristics of individual nodes, we first apply a GCN to aggregate the features of neighboring nodes within the subgraph:

\begin{equation}
\mathbf{H}_{i}^{(\ell+1)}=\sigma\left(\widetilde{\mathbf{D}}_{i}^{-\frac{1}{2}} \widetilde{\mathbf{A}}_{i} \widetilde{\mathbf{D}}_{i}^{-\frac{1}{2}} \mathbf{H}_{i}^{(\ell)} \mathbf{W}_{NS}^{(\ell)}\right),
\label{eq1}
\end{equation}
where \(\mathbf{H}_{i}^{(\ell+1)}\) and \(\mathbf{H}_{i}^{(\ell)}\) are the hidden representation matrices of the \((\ell+1)\)-th and \(\ell\)-th layers respectively, \(\widetilde{\mathbf{D}}_{i}^{-\frac{1}{2}}\) is the inverse square root of the degree matrix,
\(\widetilde{\mathbf{A}}_{i}=\mathbf{A}_{i}+\mathbf{I}\) is the adjacency matrix with self-loops for the subgraph \(\mathcal{G}_{i}\), \(\mathbf{W}_{NS}^{(\ell)}\) refers to the weight matrix of the \(\ell\)-th layer, and \(\mathbf\sigma{(\cdot)}\) is an activation function. The output representation of GCN is denoted as \(\mathbf{S}_{i}\). Since the feature vectors of the target nodes are masked, we use a Multilayer Perceptron (MLP) to process the features of the target nodes:
\begin{equation}
\boldsymbol{n}_{i}^{(\ell+1)}=\sigma\left(\boldsymbol{n}_{i}^{(\ell)} \mathbf{W}_{NS}^{(\ell)}\right),
\label{eq3}
\end{equation}
where \(\mathbf{W}_{NS}^{(\ell)}\) is shared with the GCN in Equation (\ref{eq1}). The input is the raw feature vector of the target node \(v_{i}\), and the output represents a low-dimensional embedding of \(v_{i}\). After the aggregation through GCNs, we apply average pooling to obtain a fixed-size subgraph embedding:
\begin{equation}
\boldsymbol{h}_{i}=\sum_{p=1}^{N} \frac{\left(\mathbf{S}_{i}\right)_{p}}{N},
\label{eq2}
\end{equation}
where \(\left(\mathbf{S}_{i}\right)_{p}\) is the \(p\)-th row of the subgraph embedding \(\mathbf{S}_{i}\), and \(N\) is the size of the subgraph. Subsequently, a discriminator is employed to measure the relationship between node embeddings and subgraph embeddings. This discriminator assesses the similarity between embeddings in both positive and negative instance pairs, producing a score for each. Specifically, a bilinear function is employed:
\begin{equation}
{s}_{i}^{+}={Bilinear}\left(\boldsymbol{h}_{i}, \boldsymbol{n}_{i}\right),  
{s}_{i}^{-}={Bilinear}\left(\boldsymbol{h}_{j}, \boldsymbol{n}_{i}\right),
\label{eq4}
\end{equation}
where \({s}_{i}^{+}\) is the similarity score of the positive instance pair, and \({s}_{i}^{-}\) is that of the negative instance pair formed by the target node \(\boldsymbol{n}_{i}\) and a randomly selected subgraph embedding \(\boldsymbol{h}_{j}\). Node-subgraph contrast is trained by pulling positive instance pairs closer and pushing negative instance pairs apart based on similarities~\cite{jaiswal2020survey,li2021contrastive}. This process can be seen as a binary classification task. Hence, we employ Binary Cross-Entropy (BCE) loss for training:
\begin{equation}
\mathcal{L}_{NS}=-\sum_{i=1}^{n}\left(y_{i} \log \left(s_{i}\right)+\left(1-y_{i}\right) \log \left(1-s_{i}\right)\right),
\label{eq5}
\end{equation}
where \(y_{i}\) is the label of the instance pair, \(y_{i}=1\) represents the label of positive instance pair, \(y_{i}=0\) represents the label pf negative instance pair, and \(s_{i}\) represents either \(s_{i}^{+}\) or \(s_{i}^{-}\).

\textbf{Node-Node Contrast.} Node-node contrast is more effective in identifying node-level anomalies~\cite{jin2021anemone}. Each instance pair consists of the raw node embedding and the updated node embedding, which is aggregated from neighboring nodes within the subgraph. To reduce bias, the subgraphs for negative instance pairs are the same as those sampled in node-subgraph contrast. The aggregation of subgraph embeddings is accomplished through a new GCN:
\begin{equation}
\hat{\mathbf{H}_{i}}^{(\ell+1)}=\sigma\left(\widetilde{\mathbf{D}}_{i}^{-\frac{1}{2}} \widetilde{\mathbf{A}}_{i} \widetilde{\mathbf{D}}_{i}^{-\frac{1}{2}} \hat{\mathbf{H}_{i}}^{(\ell)} \mathbf{W}_{NN}^{(\ell)}\right),
\label{eq6}
\end{equation}
where \(\mathbf{W}_{NN}^{(\ell)}\) is different from \(\mathbf{W}_{SN}^{(\ell)}\). Subsequently, we utilize a new MLP to project the node feature vectors into a unified embedding space:
\begin{equation}
\hat{\boldsymbol{n}_{i}}^{(\ell+1)}=\sigma\left(\hat{\boldsymbol{n}_{i}}^{(\ell)} \mathbf{W}_{NN}^{(\ell)}\right).
\label{eq7}
\end{equation}

The updated node embedding is obtained after aggregating information from its first-order neighboring nodes:

\begin{equation}
\hat{\boldsymbol{h}_{i}}^{(\ell+1)}=\hat{\mathbf{H}_{i}}^{(\ell+1)}[1,:].
\end{equation}

Similarly, a new bilinear function is employed to evaluate the similarities of different instance pairs:
\begin{equation}
\hat{{s}_{i}}^{+}=\hat{{Bilinear}}\left(\hat{\boldsymbol{h}_{i}}, \hat{\boldsymbol{n}_{i}}\right),  \hat{{s}_{i}}^{-}=\hat{{Bilinear}}\left(\hat{\boldsymbol{h}_{j}}, \hat{\boldsymbol{n}_{i}}\right).
\label{eq8}
\end{equation}

\subsection{Progressive Purification}
\textbf{Node Contrast Score.} For normal nodes, the similarities of positive instance pairs converge to 1, while those of negative instance pairs approach 0. In contrast, for anomalous nodes, these scores are much closer. Thus, we assess the anomaly degree of nodes by examining the difference in similarities between positive and negative instance pairs. Specifically, we subtract the similarity score of the positive pair from the negative one, both derived from that of the anomalous graph. A larger difference signifies a higher likelihood of an anomaly:
\begin{equation}
{score}_{i}^{NS}={s}_{i,ano}^{-}-{s}_{i,ano}^{+},  
{score}_{i}^{NN}=\hat{{s}}_{i,ano}^{-}-\hat{{s}}_{i,ano}^{+}.
\label{eq9}
\end{equation}

To comprehensively evaluate anomalies across different scales, we aggregate the scores from each scale through a weighted summation approach:
\begin{equation}
{score}_{i}^{msc}=\beta {score}_{i}^{NS}+(1-\beta) {score}_{i}^{NN},
\label{eq10}
\end{equation}
where \(\beta\) is the balance parameter for different scales.

\textbf{Interfering Edges Detection.} Interfering edges significantly increase the dissimilarity between the target nodes and their local subgraphs, narrowing the score gap between positive and negative instance pairs. This hampers the contrastive learning module's ability to distinguish instance pairs, thereby undermining the model's effective training. To mitigate this issue, our first step is to accurately quantify the interference level of each individual edge. 

Interfering edges can be classified into two categories: (1) anomalous connections formed between previously unlinked normal nodes, and (2) pre-existing edges that link nodes with anomalous features. Both categories represent connections between inherently dissimilar entities, leading to deviations in expected relationships. Consequently, a reduction in a node's contrast score can be attributed to its association with a dissimilar node. If an edge connects two nodes with high contrast scores, it indicates that both nodes interact with dissimilar neighbors. This dissimilarity increases the likelihood of interference between them. Based on this assumption, we aggregate the scores of connected nodes to quantify the interference level associated with edges. Through this approach, we obtain an interference-sensitive edge detection matrix:
\begin{equation}
{E}_{i, j}=\left\{\begin{array}{ll}
{score}_{i}^{msc}+{score}_{j}^{msc}, & \mathbf{A}_{i, j}=1 \\
0, & \text { otherwise }
\end{array}\right.
.
\label{eq11}
\end{equation}

Based on the edge interference scores, we employ a ranking mechanism to sort the edges in accordance with their interference scores. This mechanism identifies and targets the top-\textit{K} edges, which possess the highest potential to undermine the contrastive learning process. These edges usually connect nodes that display significant deviations in feature distribution, thereby altering the true similarity between the nodes and subgraphs. By removing these targeted edges, we obtain a cleaner graph, which effectively reduces the impact of interfering edges on subsequent training.

\textbf{Iterative Edge Removal.} 
A one-step edge removal approach is susceptible to the influence of interfering edges, particularly during the initial phase of training. This can lead to inaccuracies in node scoring and errors in edge assessments, potentially resulting in the unintentional removal of normal edges and compromising the graph's structural integrity. To address this, we adopt an iterative approach for removing interfering edges. In each iteration, only a small portion of edges is removed. After the removal, the model’s parameters are reset to prevent biases from the previously removed edges. The model then undergoes another round of training for multi-scale anomaly awareness, recalculating node scores to better identify edges with high interference. By continuously evaluating and removing interfering edges, we minimize their impact on the score calculation, progressively enhancing the precision of the detection process.

\subsection{Score Calculation}
\textbf{Loss Function.} We perform multi-scale anomaly awareness on both the anomalous graph and the clean graph. The contrastive learning loss function is defined as follows:
\begin{equation}
\mathcal{L}=\beta(\alpha \mathcal{L}_{NS}^{ano}+ (1-\alpha) \mathcal{L}_{NS}^{cla})+(1-\beta)(\alpha \mathcal{L}_{NN}^{ano}+ (1-\alpha) \mathcal{L}_{NN}^{cla}).
\label{eq12}
\end{equation}

\textbf{Multi-Scale Contrast Score.} The clean graph can obscure genuine anomaly signals and disrupt accurate anomaly detection. Therefore, we compute scores only from the anomalous graph during the score calculation process. Additionally, due to the inherent randomness in sampling, structural anomalies might occasionally connect with normal neighbors, which may result in missed anomalies. To address this, we perform multiple rounds of sampling and detection:
\begin{equation} {score}_{i}^{msc}=\overline{{score}_{i}^{msc}}+\sqrt{\frac{1}{R} \sum_{r=1}^{R}\left({score}_{i}^{msc(r)}-\overline{{score}_{i}^{msc}}\right)^{2}},
\label{eq15}
\end{equation}
where \(\overline{{score}_{i}^{msc}}=\frac{1}{R} \sum_{r=1}^{R} {score}_{i}^{msc(r)}\), and \(R\) is the total round of anomaly detection.

\textbf{Detection Score.} When an edge is identified as interfering, it increases the likelihood that the connected nodes are anomalies as well. Building on this insight, we propose a novel detection scoring approach. We initialize a detection array with all elements set to zero. For each node that serves as an endpoint of an interfering edge, we increment the score of the corresponding node. During the score calculation process, the detection score \({score}_{i}^{dec}\) is calculated iteratively, with its value directly proportional to the frequency at which the node interacts with interfering edges. A higher score reflects a stronger correlation between the node and anomalous behavior. Finally, the overall node score is computed by aggregating the multi-scale contrast scores and the detection scores:

\begin{equation}
f{(v_i)}=\gamma {{score}_{i}^{msc}}+(1-\gamma) {{score}_{i}^{dec}},
\label{eq16}
\end{equation}
where \(\gamma\) is the balance parameter for different scores.

\subsection{Complexity Analysis}

The time complexity of RWR for a node is $\mathcal{O}(N\eta)$, where $N$ is the number of nodes in the subgraph and $\eta$ is the average degree of the graph. The time complexity of contrastive learning is $\mathcal{O}(Led+LNd^{2})$, where $L$ is the number of GCN layers, $e$ is the number of edges in the subgraph and $d$ is the feature dimension. The time complexity for edge removal is $\mathcal{O}(c + c\log w)$, where $c$ is the number of edges in the graph and $w$ is the number of edges to be removed. The overall time complexity of CVGAD is $\mathcal{O}(n(N\eta + Led + LNd^{2})(T_e + T_r + r) + m(c + c\log w))$, where $T_e + T_r$ is the number of training epochs, $r$ is the score calculation epochs and $m$ is the number of iterations. 

\section{Experiments}
\subsection{Experimental Setup}
\textbf{Datasets.} We conduct experiments on five benchmark datasets: Cora, Citeseer, PubMed~\cite{sen2008collective}, Citation and ACM~\cite{Tang2008ArnetMinerEA}. Since these datasets do not inherently contain anomalous information, we inject anomalies artificially. Following the approach proposed by~\cite{ding2019deep,liu2021anomaly,jin2021anemone}, we inject both structural anomalies and contextual anomalies into each dataset. Specifically, to inject contextual anomalies, we randomly select $n\text{'}$ anomaly nodes. For each node, we replace its feature with the attribute that exhibits the greatest difference. For structural anomalies, we randomly select 15 nodes and connect them in pairs to form a clique. This process is executed repeatedly for $m\text{'}$ times. Both types of anomalies are injected in equal numbers.

\textbf{Baselines.} In our experiment, we select seven state-of-the-art models as baselines. HCM \cite{huang2022hop} determines anomalies based on the minimum hop count between nodes. CoLA \cite{liu2021anomaly}, ANEMONE \cite{jin2021anemone}, GRADATE \cite{duan2023graph}, and NLGAD \cite{duan2023normality} are four anomaly detection models based on contrastive learning, which facilitate comparisons across different scales. Sub-CR \cite{zhang2022reconstruction} and SL-GAD \cite{zheng2021generative} combine attribute reconstruction and contrastive learning to comprehensively assess the anomalous level of nodes. To evaluate the effectiveness, we adopt ROC-AUC as the evaluation metric. 

\textbf{Parameter Settings.} In CVGAD, we employ a one-layer GCN to aggregate information from subgraphs, with both subgraph embeddings and node embeddings mapped to 64-dimensional vectors. The size of the subgraph is set to 4, and the learning rate remains fixed at 0.001. Additionally, we set the value of \(\gamma\) to 0.8. Five iterations are performed on all datasets. Specifically, for Cora, CiteSeer, and PubMed, we perform edge removal every 100 epochs, conducting a total of 500 epochs. For Citation and ACM, we perform 1000 epochs of edge removal. In the refine training phase, we conduct 200 epochs on Cora, CiteSeer, and PubMed, and 400 epochs on Citation and ACM. Besides, we implement 300 rounds of score calculation in all datasets. In addition, we set K to 0.015 for ACM and 0.01 for other datasets.


\subsection{Anomaly Detection Results}

To assess the model's anomaly detection performance, we compare CVGAD with seven baseline models on five datasets. Based on the results shown in Table \ref{detection results}, we can draw the following conclusions: (1) CVGAD achieves notable AUC gains of 1.91\%, 0.31\%, 1.20\%, 1.73\%, and 0.35\% on Cora, Citeseer, Citation, ACM, and PubMed, respectively, which demonstrates the effectiveness of the proposed model; (2) The performance of contrastive learning-based methods surpasses that of HCM. The main reason is that lower similarity is typically observed between abnormal nodes and local subgraphs. Consequently, by comparing the similarities between nodes and subgraphs, we can effectively identify anomalous nodes; (3) Sub-CR exhibits good performance, primarily because other baselines rely on a single strategy, while Sub-CR combines both generative and contrastive learning strategies. However, by mitigating the influence of interfering edges, CVGAD achieves better results than Sub-CR on all datasets. 

\begin{table}
\small
  \caption{AUC results on five datasets with injected anomalies. The best results are highlighted in bold and the second-best results are underlined.}
  \label{detection results}
    \centering
  \begin{tabular}{cccccccc}
    \toprule   
    Methods     & Cora     & Citeseer   & Citation  & ACM & PubMed\\
    \midrule
    HCM  & 0.6057  & 0.6620    & 0.5624 & 0.5722  & 0.8143\\
    CoLA  & 0.9036  &  0.8878  & 0.7532 & 0.7937 & 0.9497\\
    ANEMONE & 0.9109  & 0.9252   & 0.7868 & 0.8081  & 0.9503\\
    SL-GAD  & 0.9098	& 0.9231	& 0.7824	& \underline{0.8145} & 0.9609 \\
    Sub-CR  & 0.9058  & 0.9334    & \underline{0.7973} & 0.8128  & \underline{0.9644}\\
    GRADATE  & 0.9010  &  0.8859   & 0.7414 & 0.7508 & 0.9504 \\
    NLGAD  & \underline{0.9148}  & \underline{0.9391}    & 0.7752 & 0.8065 
 & 0.9612\\
    \midrule
    CVGAD  & \textbf{0.9339}  & \textbf{0.9422}    & \textbf{0.8093} & \textbf{0.8318} & \textbf{0.9679}\\
    \bottomrule
  \end{tabular}
\end{table}

\subsection{Ablation Study}

\begin{table}
\small
  \caption{AUC results of CVGAD and its variants.}
  \label{ablation study}
   \centering
  \begin{tabular}{lcccccc} 
    \toprule   
          & Cora     & Citeseer   & Citation  & ACM & PubMed\\
    \midrule
    CVGAD\(^{sim}\)  & 0.9188  & 0.9213   & 0.7972 & 0.8067 & 0.9526\\
    CVGAD\(^{ore}\)  & 0.9194  & 0.9252    & 0.8023 & 0.8249 & 0.9587\\
    CVGAD\(^{bano}\)  & 0.9167  & 0.9199   & 0.7832 & 0.8167 & 0.9551\\
    CVGAD\(^{bcla}\)  & 0.9246  & 0.9323  & 0.8031 & 0.8274 & 0.9605 \\
    CVGAD\(^{ocla}\)  & 0.9233  & 0.9308    & 0.7997 & 0.8263 & 0.9585\\
    CVGAD\(^{con}\)  & 0.9231  & 0.9346    & 0.8011 & 0.8268 & 0.9566\\
    \midrule
    CVGAD   & \textbf{0.9339}  & \textbf{0.9422}   & \textbf{0.8093} & \textbf{0.8318} & \textbf{0.9679}\\
    \bottomrule
  \end{tabular}
\end{table}
To verify the effectiveness of different modules, we conduct three types of ablation studies. First, we examine the effect of edge removal strategies. Next, we test the impact of random walks on different views. Finally, we compare the multi-score integration approach with scores solely based on contrastive learning. The results are presented in Table \ref{ablation study}, and the detailed analysis is as follows.

\textbf{Edge Removal and Iterative Strategy.} CVGAD\(^{sim}\) performs edge removal based on the similarity scores of raw features (maintaining the same edge removal ratio as CVGAD). CVGAD\(^{ore}\) employs a non-iterative edge removal by removing the same total proportion at one step. CVGAD\(^{bano}\) trains on two anomalous graphs. The experimental results show that CVGAD exhibits superior performance over the other three models. First, it demonstrates that removing edges can effectively improve detection performance. Moreover, it indicates that the interfering edge removal method of CVGAD outperforms the approach based on similarity. Edges with low similarity are not necessarily interfering edges. Mistakenly removing normal edges can disrupt the original topology of the graph, thereby compromising the effectiveness of the detection. Additionally, the results validate that iterative edge removal is more effective than non-iterative edge removal.  This is because a one-step removal approach may be susceptible to interfering edges in the initial stage, which hinders the identification of anomalous nodes and interfering edges. In contrast, the iterative process in CVGAD gradually removes small portions of edges in each round, progressively refining the graph while simultaneously preserving the normal patterns. This approach creates a synergistic effect between the multi-scale anomaly awareness module and the progressive purification module, leading to improved performance.

\textbf{Multi-View RWR Strategy.} CVGAD\(^{bano}\) and CVGAD\(^{bcla}\) modules conduct RWR on two anomalous graphs and two clean graphs, respectively. CVGAD\(^{ocla}\) traverses only on the clean graph. The results demonstrate that random walks on two views (the clean graph and the anomalous graph) yield better performance. By conducting random walks on the clean graph, sampling interfering edges is avoided, which enhances the similarities between nodes and their neighborhood subgraphs. This improves the effectiveness of the contrastive learning process. On the other hand, performing random walks on the anomalous graph helps the model learn diverse features.
 
\textbf{Multi-Score Integration Strategy.} The score of CVGAD\(^{con}\) is determined solely by the node contrast scores. The result reveals that the combination scores yield superior performance, as nodes connected by high-interference edges are more likely to be anomalous. Detection scores provide a more precise differentiation of abnormality levels.

\subsection{Parameter Analysis}
\begin{figure}[h]
\centering  
\subfloat[Citeseer]{   
\includegraphics[scale=0.161]{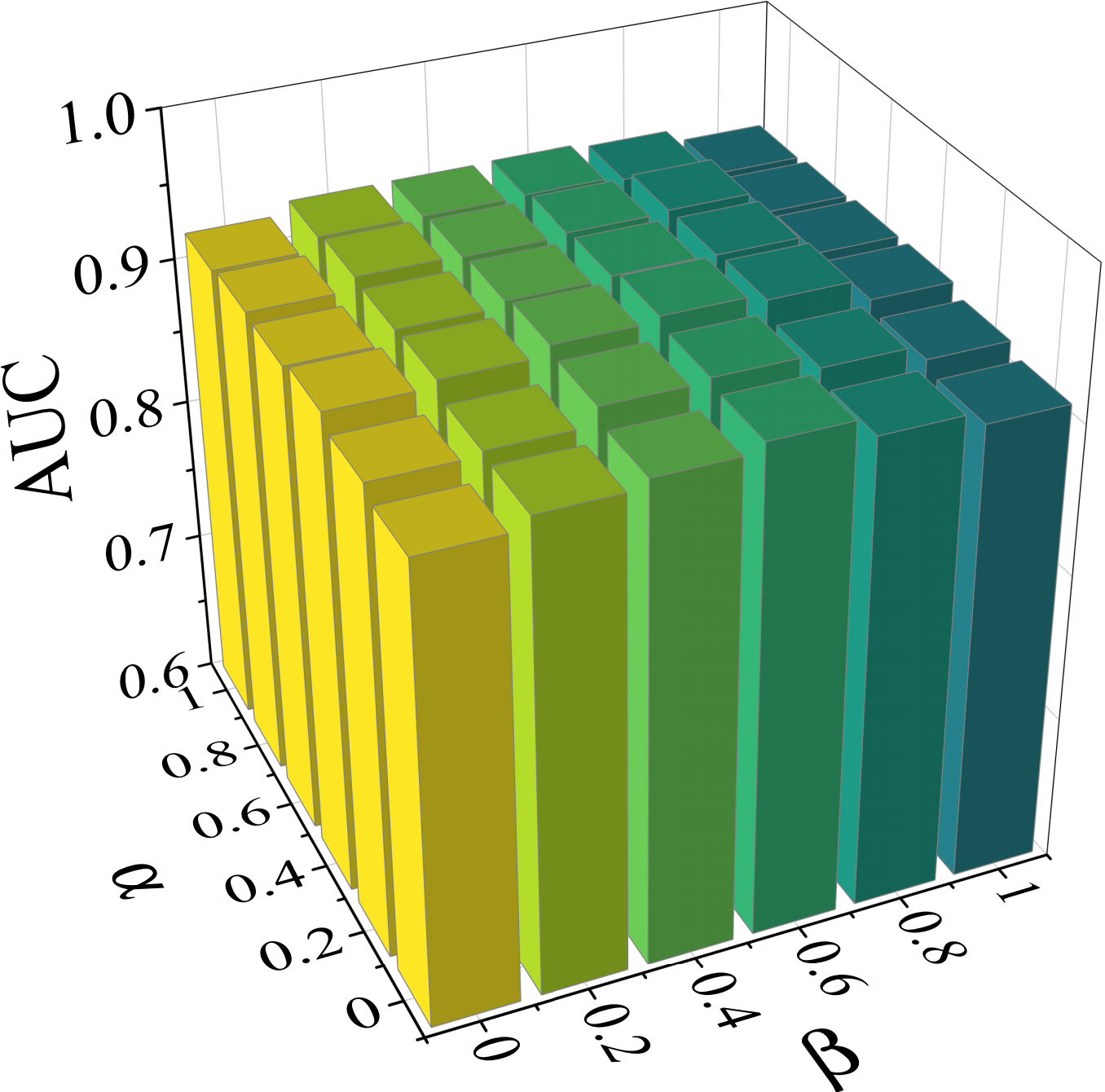}  
}
\subfloat[Citation]{ 
\includegraphics[scale=0.161]{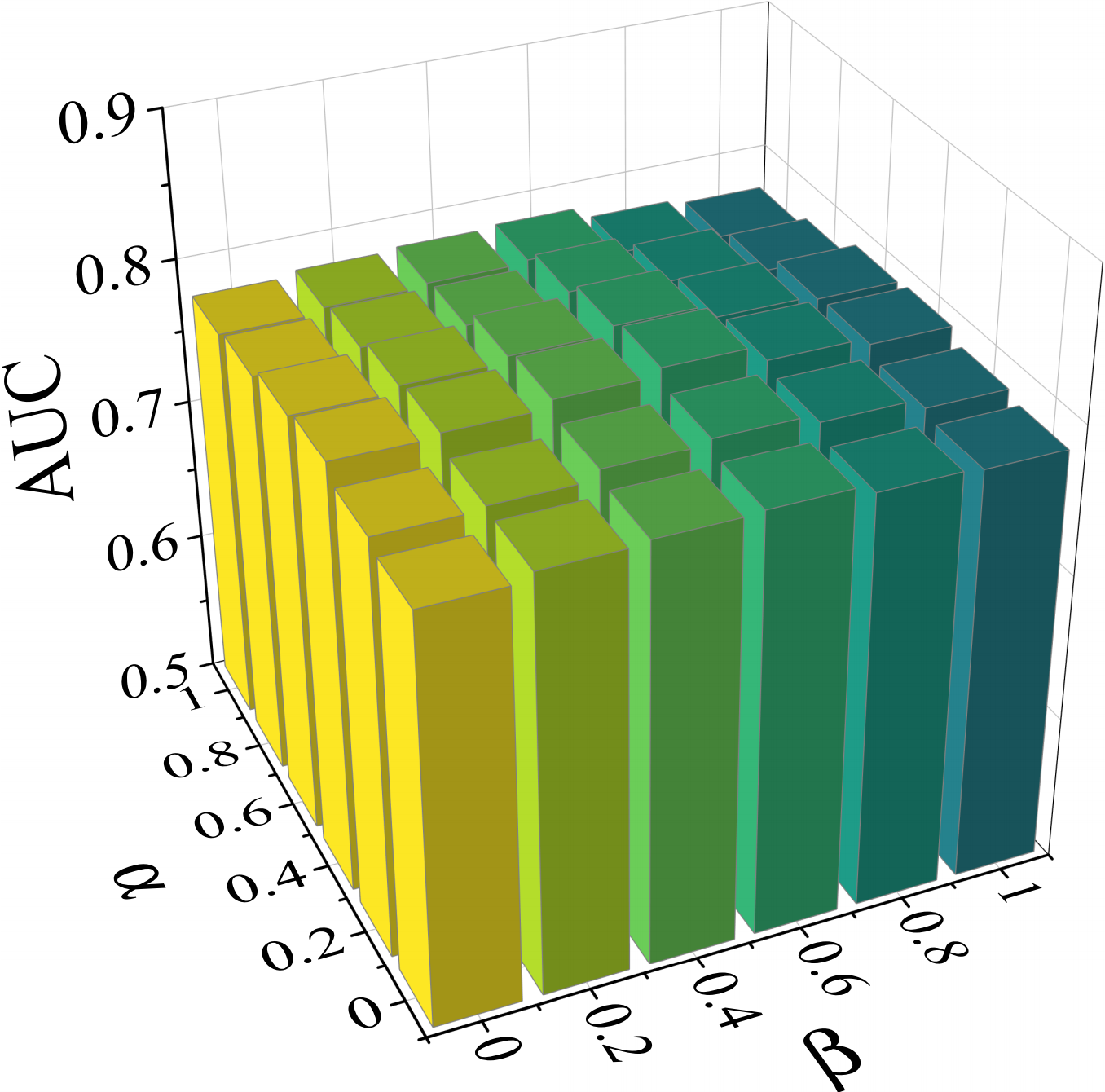}
}
\caption{Sensibility analysis of balance parameters.}    
\label{fig:alpha1}    
\end{figure}

\textbf{Balance Parameter \(\alpha\) and \(\beta\).} We investigate the impact of the view balance parameter \(\alpha\) and the scale balance parameter \(\beta\). As shown in Figure \ref{fig:alpha1}(a) and Figure \ref{fig:alpha1}(b), the performance initially increases and subsequently exhibits a decline on both the \textit{x}-axis and \textit{y}-axis. Other datasets exhibit the same phenomenon. This illustrates that employing multi-scale learning and multi-view training yields superior results compared to singular methods. The parameter \(\alpha\) controls the balance between different views. For most datasets, the optimal value of \(\alpha\) is relatively low, highlighting the effectiveness of removing interfering edges. Based on these results, we set \(\alpha\) to 0.8, 0.4, 0.4, 0.6, and 0.4 for Cora, Citeseer, Citation, ACM, and PubMed, respectively. Regarding the scale balance parameter \(\beta\), relying solely on the first-order neighbor similarity may fail to capture the full range of anomalous patterns in all datasets. In contrast, NS contrast can better reflect the broader anomalous patterns of the nodes. So we set the value of \(\beta\) at 0.6 on all datasets.

\subsection{Edge Removal Accuracy}
\label{edge-removal}
\begin{figure}[h]
\subfloat[Cora]{   
\centering 
\includegraphics[scale=0.24]{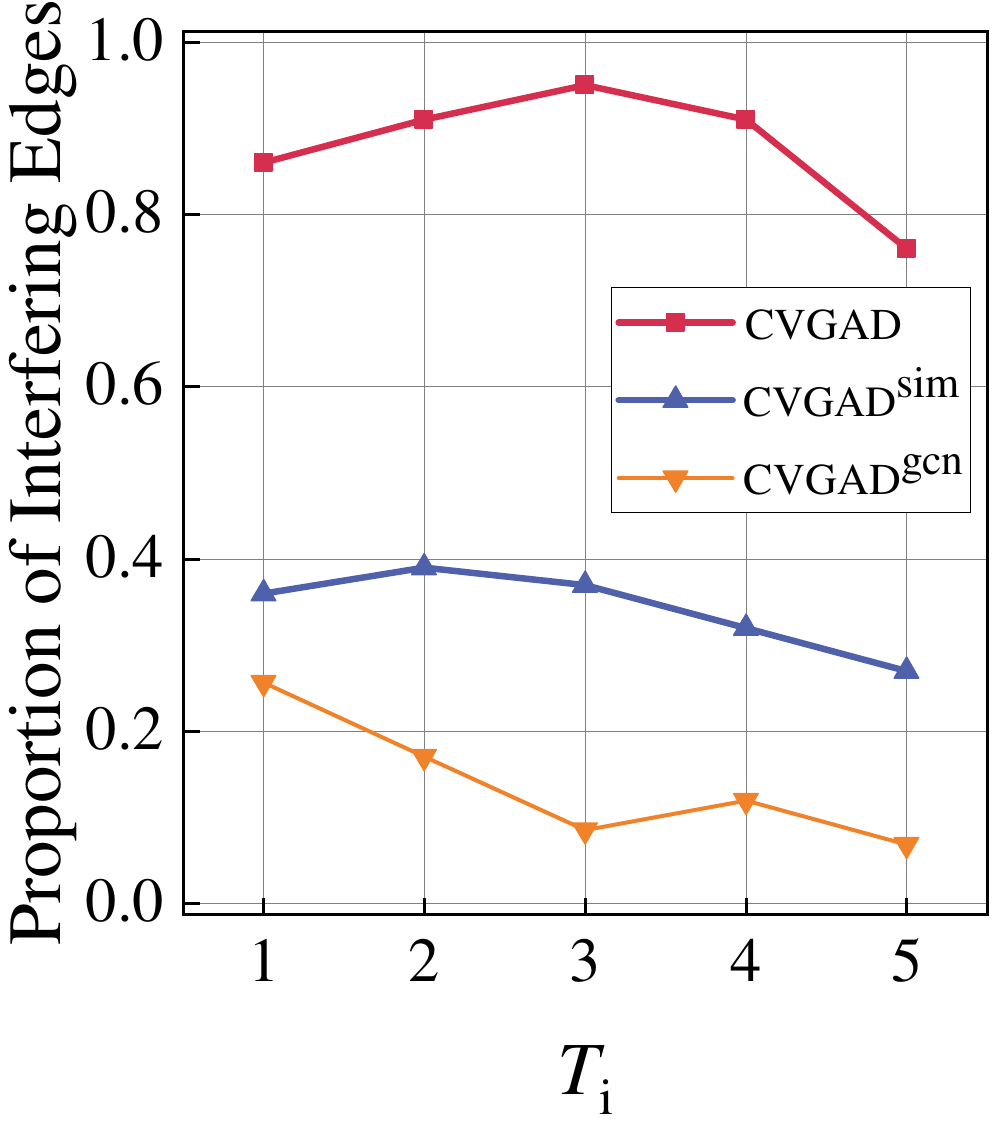}  
}
\centering 
\subfloat[ACM]{ 
\includegraphics[scale=0.24]{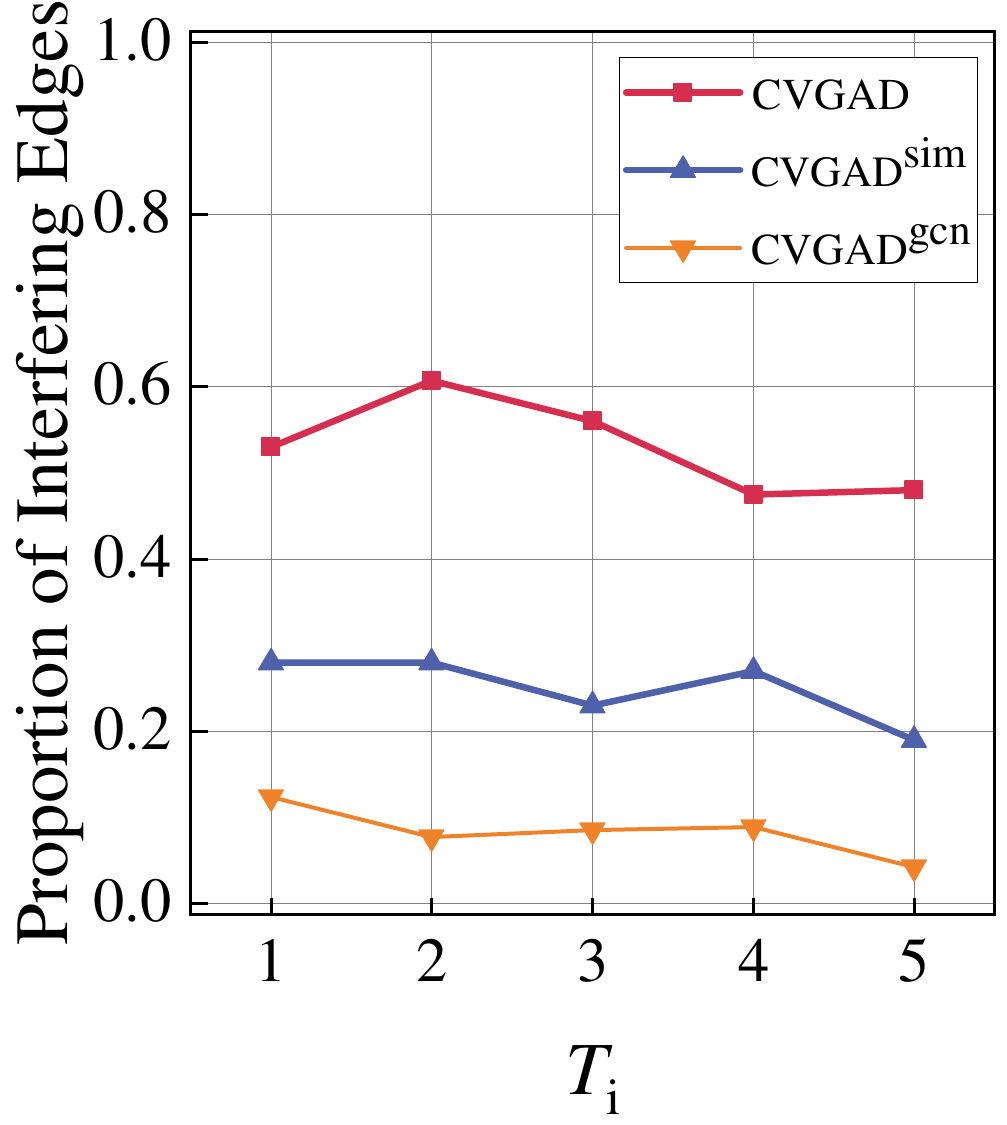}
}
\caption{Proportion of interfering edges removed.}    
\label{fig:removal1}    
\end{figure}

To validate the accuracy of the edge removal method, we compare CVGAD with edge removal methods based on raw feature similarity (CVGAD\(^{sim}\)) and GCN-aggregated feature similarity (CVGAD\(^{gcn}\)). For CVGAD, we record the proportion of interfering edges removed at each iteration. For CVGAD\(^{sim}\) and CVGAD\(^{gcn}\), we select the edges with the top-\textit{K} lowest scores and record the proportion. If the proportion of edges with zero feature similarity exceeds the top-\textit{K} threshold, we randomly select a set of top-\textit{K} edges for removal. As shown in Figure \ref{fig:removal1}, CVGAD shows significantly superior edge removal accuracy in comparison to the other methods on the Cora and ACM datasets. The reason is that CVGAD effectively integrates feature information and graph structures to distinguish interfering edges. In contrast, CVGAD\(^{sim}\) focuses solely on feature-level information, and the convolutional operation of CVGAD\(^{gcn}\) smooths out anomalous signals, which impairs the model's ability to identify anomalies. Additionally, owing to the higher heterogeneity of ACM compared to Cora, the edge identification accuracy of CVGAD is lower on ACM. 

\section{Conclusion}
We propose a novel graph anomaly detection method named CVGAD, which introduces an effective solution to the limitation of existing anomaly detection methods based on contrastive learning. CVGAD evaluates interference scores for edges and removes interfering edges to enhance the similarities between nodes and their local subgraphs. Through an iterative strategy, multi-scale anomaly awareness and progressive purification mutually reinforce each other, thereby leading to improved anomaly detection performance. Extensive experiments demonstrate the effectiveness of our approach.

\section{Acknowledgements}
This work was supported by the National Key Research
and Development Program of China (2023YFC3304503),
the National Natural Science Foundation of China (No. 62302333, 92370111, 62422210, 62272340, 62276187), and Hebei Natural Science Foundation (No. F2024202047).

\bibliographystyle{named}
\bibliography{ijcai25}

\begin{thebibliography}{}

\bibitem[\protect\citeauthoryear{Aghabozorgi and Khayyambashi}{2018}]{Aghabozorgi2018ANS}
Farshad Aghabozorgi and Mohammad~Reza Khayyambashi.
\newblock A new similarity measure for link prediction based on local structures in social networks.
\newblock {\em Physica A: Statistical Mechanics and its Applications}, 501:12--23, 2018.

\bibitem[\protect\citeauthoryear{Ding \bgroup \em et al.\egroup }{2019}]{ding2019deep}
Kaize Ding, Jundong Li, Rohit Bhanushali, and Huan Liu.
\newblock Deep anomaly detection on attributed networks.
\newblock In {\em Proceedings of the 2019 {SIAM} International Conference on Data Mining}, pages 594--602, 2019.

\bibitem[\protect\citeauthoryear{Du \bgroup \em et al.\egroup }{2023}]{du2023superdisco}
Yingjun Du, Jiayi Shen, Xiantong Zhen, and Cees~GM Snoek.
\newblock Superdisco: Super-class discovery improves visual recognition for the long-tail.
\newblock In {\em Proceedings of the IEEE Conference on Computer Vision and Pattern Recognition}, pages 19944--19954, 2023.

\bibitem[\protect\citeauthoryear{Duan \bgroup \em et al.\egroup }{2023a}]{duan2023graph}
Jingcan Duan, Siwei Wang, Pei Zhang, En~Zhu, Jingtao Hu, Hu~Jin, Yue Liu, and Zhibin Dong.
\newblock Graph anomaly detection via multi-scale contrastive learning networks with augmented view.
\newblock In {\em Proceedings of the Thirty-Seventh {AAAI} Conference on Artificial Intelligence}, pages 7459--7467, 2023.

\bibitem[\protect\citeauthoryear{Duan \bgroup \em et al.\egroup }{2023b}]{duan2023normality}
Jingcan Duan, Pei Zhang, Siwei Wang, Jingtao Hu, Hu~Jin, Jiaxin Zhang, Haifang Zhou, and Xinwang Liu.
\newblock Normality learning-based graph anomaly detection via multi-scale contrastive learning.
\newblock In {\em Proceedings of the Thirty-First ACM International Conference on Multimedia}, pages 7502--7511, 2023.

\bibitem[\protect\citeauthoryear{Fan \bgroup \em et al.\egroup }{2020}]{fan2020anomalydae}
Haoyi Fan, Fengbin Zhang, and Zuoyong Li.
\newblock Anomalydae: Dual autoencoder for anomaly detection on attributed networks.
\newblock In {\em Proceedings of the 2020 {IEEE} International Conference on Acoustics, Speech and Signal Processing}, pages 5685--5689, 2020.

\bibitem[\protect\citeauthoryear{Feng \bgroup \em et al.\egroup }{2024}]{feng2024backdoor}
Bingdao Feng, Di~Jin, Xiaobao Wang, Fangyu Cheng, and Siqi Guo.
\newblock Backdoor attacks on unsupervised graph representation learning.
\newblock {\em Neural Networks}, 180:106668, 2024.

\bibitem[\protect\citeauthoryear{Hamilton \bgroup \em et al.\egroup }{2017}]{hamilton2017inductive}
Will Hamilton, Zhitao Ying, and Jure Leskovec.
\newblock Inductive representation learning on large graphs.
\newblock In {\em Proceedings of the Annual Conference on Neural Information Processing Systems}, page 1025–1035, 2017.

\bibitem[\protect\citeauthoryear{He \bgroup \em et al.\egroup }{2024a}]{he2024tut4crs}
Dongxiao He, Jinghan Zhang, Xiaobao Wang, Meng Ge, Zhiyong Feng, Longbiao Wang, and Xiaoke Ma.
\newblock Tut4crs: Time-aware user-preference tracking for conversational recommendation system.
\newblock In {\em Proceedings of the Thirty-Second ACM International Conference on Multimedia}, pages 5856--5864, 2024.

\bibitem[\protect\citeauthoryear{He \bgroup \em et al.\egroup }{2024b}]{he2024ada}
Junwei He, Qianqian Xu, Yangbangyan Jiang, Zitai Wang, and Qingming Huang.
\newblock Ada-gad: Anomaly-denoised autoencoders for graph anomaly detection.
\newblock In {\em Proceedings of the Thirty-Eighth {AAAI} Conference on Artificial Intelligence}, pages 8481--8489, 2024.

\bibitem[\protect\citeauthoryear{Huang \bgroup \em et al.\egroup }{2022}]{huang2022hop}
Tianjin Huang, Yulong Pei, Vlado Menkovski, and Mykola Pechenizkiy.
\newblock Hop-count based self-supervised anomaly detection on attributed networks.
\newblock In {\em Proceedings of the Joint European Conference on Machine Learning and Knowledge Discovery in Databases}, pages 225--241, 2022.

\bibitem[\protect\citeauthoryear{Jaiswal \bgroup \em et al.\egroup }{2020}]{jaiswal2020survey}
Ashish Jaiswal, Ashwin~Ramesh Babu, Mohammad~Zaki Zadeh, Debapriya Banerjee, and Fillia Makedon.
\newblock A survey on contrastive self-supervised learning.
\newblock {\em Technologies}, 9(1):2, 2020.

\bibitem[\protect\citeauthoryear{Jin \bgroup \em et al.\egroup }{2021}]{jin2021anemone}
Ming Jin, Yixin Liu, Yu~Zheng, Lianhua Chi, Yuan-Fang Li, and Shirui Pan.
\newblock Anemone: Graph anomaly detection with multi-scale contrastive learning.
\newblock In {\em Proceedings of the Thirtieth ACM International Conference on Information and Knowledge Management}, pages 3122--3126, 2021.

\bibitem[\protect\citeauthoryear{Jin \bgroup \em et al.\egroup }{2023}]{jin2023local}
Di~Jin, Bingdao Feng, Siqi Guo, Xiaobao Wang, Jianguo Wei, and Zhen Wang.
\newblock Local-global defense against unsupervised adversarial attacks on graphs.
\newblock In {\em Proceedings of the Thirty-Seventh {AAAI} Conference on Artificial Intelligence}, pages 8105--8113, 2023.

\bibitem[\protect\citeauthoryear{Jin \bgroup \em et al.\egroup }{2025}]{jin2025backdoor}
Di~Jin, Yujun Zhang, Bingdao Feng, Xiaobao Wang, Dongxiao He, and Zhen Wang.
\newblock Backdoor attack on propagation-based rumor detectors.
\newblock In {\em Proceedings of the Thirty-Ninth {AAAI} Conference on Artificial Intelligence}, pages 17680--17688, 2025.

\bibitem[\protect\citeauthoryear{Kim \bgroup \em et al.\egroup }{2022}]{kim2022graph}
Hwan Kim, Byung~Suk Lee, Won-Yong Shin, and Sungsu Lim.
\newblock Graph anomaly detection with graph neural networks: Current status and challenges.
\newblock {\em IEEE Access}, 10:111820--111829, 2022.

\bibitem[\protect\citeauthoryear{Kumar \bgroup \em et al.\egroup }{2020}]{KUMAR2020124289}
Ajay Kumar, Shashank~Sheshar Singh, Kuldeep Singh, and Bhaskar Biswas.
\newblock Link prediction techniques, applications, and performance: A survey.
\newblock {\em Physica A: Statistical Mechanics and its Applications}, 553:124289, 2020.

\bibitem[\protect\citeauthoryear{Lanciano \bgroup \em et al.\egroup }{2020}]{Lanciano2020ExplainableCO}
Tommaso Lanciano, Francesco Bonchi, and A.~Gionis.
\newblock Explainable classification of brain networks via contrast subgraphs.
\newblock In {\em Proceedings of the Twenty-Sixth ACM SIGKDD Conference on Knowledge Discovery and Data Mining}, pages 3308--3318, 2020.

\bibitem[\protect\citeauthoryear{Li \bgroup \em et al.\egroup }{2017}]{li2017radar}
Jundong Li, Harsh Dani, Xia Hu, and Huan Liu.
\newblock Radar: Residual analysis for anomaly detection in attributed networks.
\newblock In {\em Proceedings of the Twenty-Sixth International Joint Conference on Artificial Intelligence}, pages 2152--2158, 2017.

\bibitem[\protect\citeauthoryear{Li \bgroup \em et al.\egroup }{2021}]{li2021contrastive}
Yunfan Li, Peng Hu, Zitao Liu, Dezhong Peng, Joey~Tianyi Zhou, and Xi~Peng.
\newblock Contrastive clustering.
\newblock In {\em Proceedings of the Thirty-Fifth {AAAI} Conference on Artificial Intelligence}, pages 8547--8555, 2021.

\bibitem[\protect\citeauthoryear{Li \bgroup \em et al.\egroup }{2024}]{li2024diffgad}
Jinghan Li, Yuan Gao, Jinda Lu, Junfeng Fang, Congcong Wen, Hui Lin, and Xiang Wang.
\newblock Diffgad: A diffusion-based unsupervised graph anomaly detector.
\newblock {\em Proceedings of the International Conference on Learning Representations}, 2024.

\bibitem[\protect\citeauthoryear{Liu \bgroup \em et al.\egroup }{2021}]{liu2021anomaly}
Yixin Liu, Zhao Li, Shirui Pan, Chen Gong, Chuan Zhou, and George Karypis.
\newblock Anomaly detection on attributed networks via contrastive self-supervised learning.
\newblock {\em IEEE Transactions on Neural Networks and Learning Systems}, 33(6):2378--2392, 2021.

\bibitem[\protect\citeauthoryear{Liu \bgroup \em et al.\egroup }{2024}]{liu2024hierarchical}
Yuting Liu, Liu Yang, and Yu~Wang.
\newblock Hierarchical fine-grained visual classification leveraging consistent hierarchical knowledge.
\newblock In {\em Proceedings of the Joint European Conference on Machine Learning and Knowledge Discovery in Databases}, pages 279--295, 2024.

\bibitem[\protect\citeauthoryear{Luo \bgroup \em et al.\egroup }{2022}]{luo2022comga}
Xuexiong Luo, Jia Wu, Amin Beheshti, Jian Yang, Xiankun Zhang, Yuan Wang, and Shan Xue.
\newblock Comga: Community-aware attributed graph anomaly detection.
\newblock In {\em Proceedings of the Fifteenth ACM International Conference on Web Search and Data Mining}, pages 657--665, 2022.

\bibitem[\protect\citeauthoryear{Noble and Cook}{2003}]{noble2003graph}
Caleb~C Noble and Diane~J Cook.
\newblock Graph-based anomaly detection.
\newblock In {\em Proceedings of the ninth ACM SIGKDD Conference on Knowledge Discovery and Data Mining}, pages 631--636, 2003.

\bibitem[\protect\citeauthoryear{Peng \bgroup \em et al.\egroup }{2018}]{peng2018anomalous}
Zhen Peng, Minnan Luo, Jundong Li, Huan Liu, Qinghua Zheng, et~al.
\newblock Anomalous: A joint modeling approach for anomaly detection on attributed networks.
\newblock In {\em Proceedings of the Twenty-Seventh International Joint Conference on Artificial Intelligence}, pages 3513--3519, 2018.

\bibitem[\protect\citeauthoryear{Perozzi and Akoglu}{2016}]{perozzi2016scalable}
Bryan Perozzi and Leman Akoglu.
\newblock Scalable anomaly ranking of attributed neighborhoods.
\newblock In {\em Proceedings of the 2016 SIAM International Conference on Data Mining}, pages 207--215, 2016.

\bibitem[\protect\citeauthoryear{Sen \bgroup \em et al.\egroup }{2008}]{sen2008collective}
Prithviraj Sen, Galileo Namata, Mustafa Bilgic, Lise Getoor, Brian Galligher, and Tina Eliassi-Rad.
\newblock Collective classification in network data.
\newblock {\em AI Magazine}, 29(3):93--93, 2008.

\bibitem[\protect\citeauthoryear{Tang \bgroup \em et al.\egroup }{2008}]{Tang2008ArnetMinerEA}
Jie Tang, Jing Zhang, Limin Yao, Juan-Zi Li, Li~Zhang, and Zhong Su.
\newblock Arnetminer: extraction and mining of academic social networks.
\newblock In {\em Proceedings of the Fourteenth ACM SIGKDD Conference on Knowledge Discovery and Data Mining}, pages 990--998, 2008.

\bibitem[\protect\citeauthoryear{Tong \bgroup \em et al.\egroup }{2006}]{tong2006fast}
Hanghang Tong, Christos Faloutsos, and Jia-Yu Pan.
\newblock Fast random walk with restart and its applications.
\newblock In {\em Proceedings of the Sixth {IEEE} International Conference on Data Mining}, pages 613--622, 2006.

\bibitem[\protect\citeauthoryear{Wang \bgroup \em et al.\egroup }{2023}]{wang2023augmenting}
Xiaobao Wang, Yiqi Dong, Di~Jin, Yawen Li, Longbiao Wang, and Jianwu Dang.
\newblock Augmenting affective dependency graph via iterative incongruity graph learning for sarcasm detection.
\newblock In {\em Proceedings of the Thirty-Seventh {AAAI} Conference on Artificial Intelligence}, pages 4702--4710, 2023.

\bibitem[\protect\citeauthoryear{Wang \bgroup \em et al.\egroup }{2025}]{wang2025elevating}
Xiaobao Wang, Yujing Wang, Dongxiao He, Zhe Yu, Yawen Li, Longbiao Wang, Jianwu Dang, and Di~Jin.
\newblock Elevating knowledge-enhanced entity and relationship understanding for sarcasm detection.
\newblock {\em IEEE Transactions on Knowledge and Data Engineering}, 2025.

\bibitem[\protect\citeauthoryear{Wu \bgroup \em et al.\egroup }{2019}]{wu2019session}
Shu Wu, Yuyuan Tang, Yanqiao Zhu, Liang Wang, Xing Xie, and Tieniu Tan.
\newblock Session-based recommendation with graph neural networks.
\newblock In {\em Proceedings of the Thirty-Third {AAAI} Conference on Artificial Intelligence}, pages 346--353, 2019.

\bibitem[\protect\citeauthoryear{Xiang \bgroup \em et al.\egroup }{2023}]{XiangZHQDDB023}
Haolong Xiang, Xuyun Zhang, Hongsheng Hu, Lianyong Qi, Wanchun Dou, Mark Dras, Amin Beheshti, and Xiaolong Xu.
\newblock Optiforest: Optimal isolation forest for anomaly detection.
\newblock In {\em Proceedings of the Thirty-Second International Joint Conference on Artificial Intelligence}, pages 2379--2387, 2023.

\bibitem[\protect\citeauthoryear{Xiang \bgroup \em et al.\egroup }{2024}]{xiang2024federated}
Haolong Xiang, Xuyun Zhang, Xiaolong Xu, Amin Beheshti, Lianyong Qi, Yujie Hong, and Wanchun Dou.
\newblock Federated learning-based anomaly detection with isolation forest in the iot-edge continuum.
\newblock {\em ACM Transactions on Multimedia Computing, Communications and Applications}, 2024.

\bibitem[\protect\citeauthoryear{Zhang \bgroup \em et al.\egroup }{2022}]{zhang2022reconstruction}
Jiaqiang Zhang, Senzhang Wang, and Songcan Chen.
\newblock Reconstruction enhanced multi-view contrastive learning for anomaly detection on attributed networks.
\newblock In {\em Proceedings of the Thirty-First International Joint Conference on Artificial Intelligence}, pages 2376--2382, 2022.

\bibitem[\protect\citeauthoryear{Zheng \bgroup \em et al.\egroup }{2021}]{zheng2021generative}
Yu~Zheng, Ming Jin, Yixin Liu, Lianhua Chi, Khoa~T Phan, and Yi-Ping~Phoebe Chen.
\newblock Generative and contrastive self-supervised learning for graph anomaly detection.
\newblock {\em IEEE Transactions on Knowledge and Data Engineering}, 35(12):12220--12233, 2021.

\bibitem[\protect\citeauthoryear{Zou \bgroup \em et al.\egroup }{2024}]{zou2024structural}
Dongcheng Zou, Hao Peng, and Chunyang Liu.
\newblock A structural information guided hierarchical reconstruction for graph anomaly detection.
\newblock In {\em Proceedings of the Thirty-Third ACM International Conference on Information and Knowledge Management}, pages 4318--4323, 2024.

\end{thebibliography}

\end{document}